\documentclass{article}
\usepackage{spconf,amsmath,graphicx}
\usepackage{multirow}

\title{accelerating recurrent neural network language model based online speech recognition system}
\name{Kyungmin Lee, Chiyoun Park, Namhoon Kim, and Jaewon Lee}
\address{DMC R\&D Center, Samsung Electronics, Seoul, Korea\\
         \small \{k.m.lee, chiyoun.park, namhoon.kim, jwonlee\}@samsung.com}

\begin{document}
\ninept
\maketitle
\begin{abstract}
This paper presents methods to accelerate recurrent neural network based language models (RNNLMs) for online speech recognition systems.
Firstly, a lossy compression of the past hidden layer outputs (history vector) with caching is introduced in order to reduce the number of LM queries.
Next, RNNLM computations are deployed in a CPU-GPU hybrid manner, which computes each layer of the model on a more advantageous platform.
The added overhead by data exchanges between CPU and GPU is compensated through a frame-wise batching strategy.
The performance of the proposed methods evaluated on LibriSpeech\footnote{http://www.openslr.org/12/} test sets indicates that the reduction in history vector precision improves the average recognition speed by 1.23 times with minimum degradation in accuracy. On the other hand, the CPU-GPU hybrid parallelization enables RNNLM based real-time recognition with a four times improvement in speed.
\end{abstract}
\section{Introduction}
\label{sec:intro}

A language model (LM) computes the likelihood of a given sentence and is used to improve the accuracy of an automatic speech recognition (ASR) system.
Recent research has focused on neural network (NN) based LMs~\cite{NNLM03:Bengio} because of their outstanding performances in generalizing from sparse data, which traditional n-gram based LMs could not do.
In particular, recurrent neural network based LMs (RNNLMs)~\cite{RNNLM10:Mikolov} do not even require Markov assumptions as they can model word histories of variable-length, and these virtues of them have helped improve the performance of many ASR systems~\cite{Example11:Kombrink,Example14:Tilk}.
However, to our knowledge, they are not yet actively adopted in real-time ASR systems due to their high computational complexities.

Several attempts have been made to utilize RNNLMs for online decoding in real-time ASR systems~\cite{WFST12:Gwnol, Ngram14:Arisoy, 2pass:Si} However, they either
simulate only some aspects of RNNLMs into the traditional architectures~\cite{WFST12:Gwnol, Ngram14:Arisoy}, or perform a 2-pass decoding~\cite{2pass:Si} which innately could not be applied before the end of the utterance was reached.
There have also been attempts to apply RNNLM directly to online ASR without approximation by eliminating redundant computations~\cite{1pass14:Hori, cache15:Lee, cache14:Huang}. In our previous research~\cite{cache15:Lee}, we were successful in applying moderate size RNNLMs directly to CPU-GPU hybrid online ASR systems with a cache strategy~\cite{cache14:Huang}.
However, in order to apply it to a more complex task with bigger RNNLMs, we needed to find a way to accelerate it further.

Recent studies indicate that one can reduce the number of distinct RNN computations by treating similar past hidden layer outputs, also referred to as history vectors, as same~\cite{lattice16:Liu}, and that RNNLMs can be accelerated with GPU parallelization~\cite{gpu14:Chen}.
In this paper, we attempt two different approaches in order to achieve real-time performance in a large RNNLM based ASR system.
Firstly, a lossy compression is applied to the cache of the history vector.
The precision of the vectors can be controlled by either rounding up with a smaller number of significant digits or at an extreme, by storing only the sign of each element.
Next, we propose GPU parallelization of RNNLM computations, but only on selected layers.
Instead of performing all RNNLM computations on the same platform, compute-intensive parts of the model are computed on GPUs,
and the parts that need to utilize a large memory are calculated on CPUs.
This method inherently increases the overhead of data transfer between CPUs and GPUs. This is handled by coordinating a batch transfer method that reduces the number of communications and the size of the data blocks at the same time in the hybrid ASR systems.

The paper is organized as follows.
The architecture of our baseline ASR system is explained in Section 2.
The lossy compression method of the history vectors is explained in Section 3.
Section 4 explains how RNNLM rescoring is accelerated with CPU-GPU parallelization.
Section 5 evaluates performance improvements of the proposed methods, followed by conclusion in Section 6.

\section{ARCHITECTURE OF OUR BASELINE CPU-GPU HYBRID RNNLM RESCORING}
\label{sec:architecture}

In the CPU-GPU hybrid ASR system~\cite{hydra12:Kim}, the weighted finite state transducer (WFST) is composed of four layers each representing an acoustic model (AM), a context model, a pronunciation model, and an LM.
WFSTs output word hypotheses when they reach word boundaries during frame-synchronous Viterbi searches and the hypotheses can be rescored by a separately stored RNNLM.
However, in order to speed up on-the-fly rescoring based on RNNLMs, we needed to reduce redundant computations as much as possible.
In this section, we briefly outline the architecture of our baseline CPU-GPU hybrid RNNLM rescoring proposed in~\cite{cache15:Lee}.
The main highlights of our baseline architecture are the use of gated recurrent unit (GRU)~\cite{DBLP:journals/corr/ChungGCB14} based RNNLM, noise contrastive estimation (NCE)~\cite{DBLP:journals/jmlr/GutmannH12} at the output layer, n-gram based maximum entropy (MaxEnt) bypass~\cite{DBLP:conf/asru/MikolovDPBC11} from input to output layers, and cache based
on-the-fly rescoring.

\subsection{GRU based RNN}
\label{ssec:gru}

\begin{figure}[t]
	\centering
	\centerline{\includegraphics[width=4.5cm]{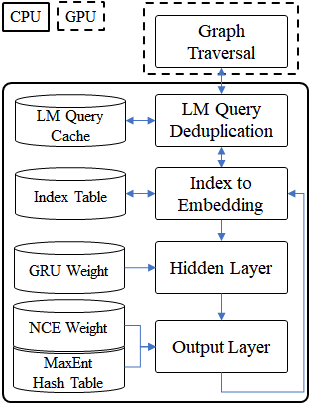}}
	\caption{On-the-fly rescoring with the LM query cache (baseline).}
	\label{fig:cpu1}
\end{figure}

We employed a GRU which is a type of gated RNNs~\cite{DBLP:journals/corr/ChungGCB14}.
The GRU is a mechanism designed to prevent vanishing gradient problems related to long-term dependencies over time by using reset gates and update gates.
For calculating a output vector of a GRU hidden layer, a total of six weight matrices and three bias vectors need to be loaded
into memory since for each gate and a candidate activation, two weight matrices and one bias vector are required.
Thus the memory usage can go up to several megabytes even if the weights are stored in a single precision floating-point format.
The computational complexities of GRU computations are $O(H \times H)$ for a hidden layer of size $H$.
This is a highly compute-intensive task considering that the number of unique LM queries in decoding an utterance can reach several hundreds of thousands.

\subsection{Noise contrastive estimation}
\label{ssec:nce}
In order to guarantee that the scores calculated at the output layer of an RNNLM are valid probabilities, they need to be normalized over different word sequences.
The normalization is a highly computationally intensive task considering the vocabulary size $V$ can reach millions.
In order to address this, we employ an NCE at the output layer~\cite{DBLP:journals/jmlr/GutmannH12}.
NCE is a sampling-based approximation method that treats partition functions as separate parameters and learns them
by non-linear logistic regression. The variances of these partition functions estimated by NCE are often limited to small values~\cite{DBLP:conf/icassp/ChenLGW15a}, allowing us to use the unnormalized scores without significant reduction in the recognition accuracy. 
Even though the only required computations are inner products between the GRU outputs and
NCE weights corresponding to the current word, the NCE weight matrix of size $H \times V$ need to be loaded into memory.

\subsection{Maximum Entropy}
\label{ssec:maxent}
The second strategy to reduce computation in our GRU based RNNLM is to use an n-gram based MaxEnt bypass connections from
input to output layers~\cite{DBLP:conf/asru/MikolovDPBC11}.
The MaxEnt scheme helps in maintaining a relatively small size for the hidden layer without significant reduction in recognition accuracy. The two types of parallel models, the main network consisting of GRUs and NCE, and the other with
MaxEnt bypass connections, operate as an ensemble model and can improve the overall recognition accuracy.
In order to reduce the computational overhead because of the bypass connections, we implemented a hash-based MaxEnt.
This method requires the loading of a large hash table proportional to the number of n-grams, to retrieve a probability for
the given n-gram in constant time.

\subsection{On-the-fly rescoring with cache}
\label{ssec:on-the-fly}

The process flow diagram of our baseline CPU-GPU hybrid RNNLM rescoring is shown in Figure~\ref{fig:cpu1}.
The LM queries with same history as well as following words are deduplicated by applying a cache strategy at the start of the rescoring procedure~\cite{cache15:Lee}.
After the deduplication, the embedding vectors corresponding to indices are retrieved by using an ``Index Table".
The RNNLM computations are then performed with appropriate values in CPU memory.
The results of the calculations are converted to indices, cached, and returned to graph traversals.

\section{QUANTIZATION OF HISTORY VECTORS}
\label{sec:quantization}

\begin{figure}[t]
	\centering
	\centerline{\includegraphics[width=4.5cm]{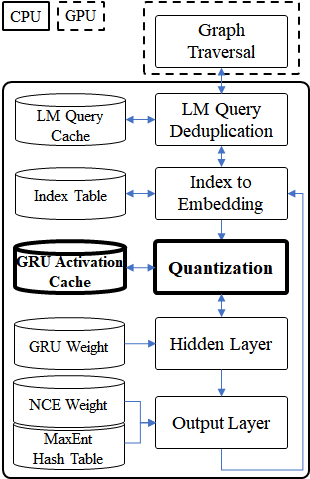}}
	\caption{Proposed on-the-fly rescoring with the cache of quantized history vectors.}
	\label{fig:cpu2}
\end{figure}

The cache-based strategy for deduplicating LM queries dramatically accelerated our baseline RNNLM rescoring with a cache hit ratio of around 89\% and more than 10 times reduction in computation time~\cite{cache15:Lee}. However, there is still room for improvement by extending this caching strategy to the outputs of GRU hidden layers.

The current GRU hidden layer outputs computed based on the previous GRU hidden layer outputs (history vectors) could be shared between similar LM queries.
Therefore, in order to reuse the precomputed history vectors,
we created another cache for that vectors just before computing RNNLMs as shown in Figure~\ref{fig:cpu2}.
The key of the cache is the GRU input which is a pair of a word embedding and a history vector,
and the value of the cache is a GRU hidden layer output corresponding to that input.
The number of unique computations is further reduced by assuming that close history vectors would result in similar GRU hidden layer outputs, with negligible effect on the overall ASR results.
Euclidean distance would be an easy way to measure the similarity~\cite{lattice16:Liu}, but it would still require a lot of computations that can slow down the whole rescoring process. Instead, we propose to quantize the history vectors by controlling the precision of history vector itself by rounding up to a specified decimal point.
We also consider an extreme case, in which we store only the signs of each element, as it would still capture some of the latent meanings which the hidden layers represent.

\begin{table}[t]
	\caption{Redundancy rates of quantized history vectors.}
	\centering
	\begin{tabular}{ccc}
		Precision & Count   & Redundancy rate \\ \hline\hline
		(baseline)  & 103,904 & 0.0   \% \\
		round-2   & 102,776 & 1.09  \% \\
		round-1   & 102,776 & 1.09  \% \\
		sign      & 88,749  & 14.59 \% \\ \hline
		
	\end{tabular}
	\label{tab:precision}
\end{table}

Table~\ref{tab:precision} shows the possible reduction of computations for a four-second utterance.
(Note that each element of the history vector ranges from -1 to 1.)
The term ``Precision" refers to the quantization of history vectors to a specified decimal place.
After the initial deduplication, in our baseline system, we have 103,904 unique LM queries as can be observed from the first row of Table~\ref{tab:precision}.
The ``round-2" row shows that only 1.09\% of the computations can be reduced by caching the history vectors rounded to the second decimal place.
Rounding off the history vectors to the first decimal place shows that there is no further redundancy.
However, as shown in the last row of Table~\ref{tab:precision}, an extreme case of quantization where only signs of each element are stored, we were able to reduce 14.59\% of the computations. 
This relatively huge reduction may affect the accuracy of RNNLM results to some extent since after the extreme sign quantization there are still $2^{256}$ possible unique history vectors for a hidden layer of size $H = 256$, but it is worthy to evaluate its effect on ASR systems. 

\section{CPU-GPU HYBRID DEPLOYMENT OF RNNLM COMPUTATION}
\label{sec:gpu}

As described in Section 2, the proposed RNNLM model cannot be readily deployed on a GPU processor due to its large memory requirement.
The word embedding step at the input layer requires space proportional to the size of vocabulary, and the MaxEnt step at the output layer need to maintain a large hash table that can store the n-grams and the corresponding scores.
Also, the NCE step at the output layer requires loading of an NCE weight matrix proportional to the size of vocabulary.
On the other hand, the hidden layer occupies only a fixed amount of memory but needs a large number of computations instead.

\begin{table}[h]
	\caption{Operation times for each RNNLM computation step in seconds.}
	\centering
	\begin{tabular}{c|ccc|c|c}
		\multirow{2}{*}{Processor} & \multicolumn{3}{c|}{Data transfer} & Hidden & Output \\ 
		& Unit & Count & Time & Layer & Layer                                \\ \hline \hline
		CPU & -       & -     & -    & 6.23 & 0.04    \\ \hline
		GPU & LM Query & 102,172 & 5.94 & 2.15 & 0.06 \\
		GPU & Frame &  518    & 0.60 & 2.26 & 0.03    \\ \hline
	\end{tabular}
	\label{tab:profile}
\end{table}

\begin{figure}[t]
	\centering
	\centerline{\includegraphics[width=8.5cm]{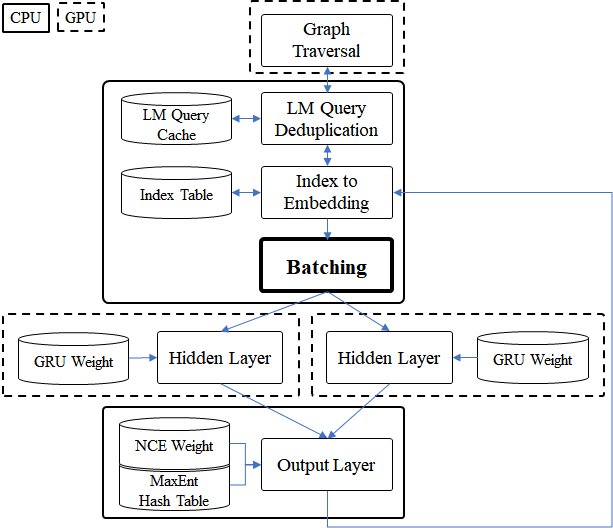}}
	\caption{Proposed GPU based RNNLM rescoring with frame-wise batch data transfer.}
	\label{fig:gpu}
\end{figure}

The first row of Table~\ref{tab:profile} shows a profiling result of an RNNLM computation with a single layer of 128 GRU nodes based on a three-second utterance.
As is expected, the hidden layer takes 99\% of the overall computation, which we aim to reduce in this section.
The high computational rates of neural networks are easily accelerated by utilization of GPUs, but their high memory requirements for word embeddings and MaxEnts prevent us from doing so.
Therefore, we deploy only the hidden layer part of the computation on the GPUs and keep the input embedding and output layer computations on the CPU side, as shown in Figure~\ref{fig:gpu}.
As can be observed from the second row of Table~\ref{tab:profile}, the hybrid deployment reduces the computation time
for the hidden layer to one-third of what was done on CPU alone.

However, this method also introduces a setback.
Because only the middle layer of the RNNLM computations was deployed on the GPU side, and its surrounding layers are computed on CPUs, the information needs to be shared across the two heterogeneous processor units frequently.
As the number of data exchanges increases, the decoding speed of the hybrid ASR system inevitably decreases.
The second row of Table~\ref{tab:profile} also shows that there have been more than a hundred thousand data exchanges during an utterance, which delayed the overall computation by 5.94 seconds, which is twice as long as the original utterance.

The frequency of data transfers between CPUs and GPUs affects the decoding speed more critically than the data size in each transfer.
Therefore, we propose a method in which we reduce the number of data copies between CPUs and GPUs by concatenating the needed
information to one block per frame. During the batching step, the history vectors and their next word embeddings that are emitted for each frame are stored in a consecutive CPU memory block, and the whole data block is transferred to GPU memory at once.
The GRU outputs from the GPU are also copied back to the output layer computation in one data block. This effect can be observed from the last row of Table~\ref{tab:profile}, in which the data transfer time is reduced to 10\% of the original.
In addition, this approach still works in multi-GPU environments without additional operations by evenly distributing the block to GPUs since the hidden layer calculations for each segment of the CPU memory block are not sequentially related
to each other.

\section{experiments}
\label{sec:experiments}
\begin{table*}[t]
	\caption{Performances on LibriSpeech's test sets; all evaluations were performed with same decoding options.}
	\centering
	\begin{tabular}{cccccccccccc}
		\hline
		\multirow{2}{*}{LM} & \multirow{2}{*}{Processor} & Rescoring & \multirow{2}{*}{Precision} & \multicolumn{2}{c}{dev-clean} & \multicolumn{2}{c}{test-clean}& \multicolumn{2}{c}{dev-other}& \multicolumn{2}{c}{test-other} \\ 
		& & threads& & WER & RTF & WER & RTF & WER & RTF & WER & RTF \\\hline \hline
		4-gram full & CPU & 4 & -   & 4.28 & 0.18 & 4.95& 0.33& 11.92& 0.54&11.87&0.26 \\ \hline
		 & \multirow{4}{*}{CPU} & \multirow{4}{*}{4} & (baseline)    & 4.05 & 2.16 & 4.69& 2.19& 11.70& 3.58&11.47&3.37 \\ 
		 &  &  & round-2    & 4.06 & 1.85 & 4.69& 1.89& 11.69& 2.91&11.49&2.85 \\     
		 &  &  & round-1    & 4.05 & 1.82 & 4.69& 1.87& 11.69& 2.95&11.48&2.89 \\		
		GRU-RNNLM &  &  & sign & 4.06 & 1.79 & 4.69& 1.80& 11.69& 2.82&11.47&2.75 \\ \cline{2-12}
		$(H = 256)$ & \multirow{4}{*}{GPU} & 1 & \multirow{4}{*}{-}    & 4.05 & 1.10 & 4.69& 1.08& 11.70& 1.94&11.49&2.29 \\
		 &  & 2 &     & 4.06 & 0.71 & 4.69& 0.70& 11.70& 1.24&11.47&1.20 \\
		 &  & 3 &     & 4.05 & 0.58 & 4.68& 0.63& 11.69& 0.98&11.47&0.97 \\
		 &  & 4 &     & 4.05 & 0.52 & 4.69& 0.52& 11.70& 0.88&11.47&0.94 \\ \hline								
	\end{tabular}
	\label{tab:performance}
\end{table*}

\subsection{Experimental setup}
\label{ssec:setup}
The LMs in our experiments were trained on the training corpus of LibriSpeech~\cite{librispeech15:Panayotov}.
To compare the performance with n-grams, ``4-gram full LM" in LibriSpeech was used.
Both vanilla-RNNLMs and GRU-RNNLMs consisted of a single hidden layer and 4-gram based MaxEnt connections.
The vocabularies used for all RNNLMs were the same as ``4-gram full LM" ($V = 200,000$).
A bi-directional recurrent deep neural network (RDNN) based AM with three hidden long short term memory (LSTM) layers (500 nodes for each layer), and a softmax output layer was trained using about 7,600 hours of the fully transcribed in-house English speech data mostly consisting of voice commands and dialogs.
WFSTs were compiled with 2-gram LMs, and all the epsilon transitions were removed so that computations on GPUs could be optimized.

The hardware specification for the evaluations was Intel Xeon E5-2680 with 12 physical CPU cores and four Nvidia Tesla K80 GPUs equipped with 12 GB memory. We used CUDA for GPU parallelization. CUBLAS, which is a linear algebra library of CUDA, was used for matrix multiplications and kernel functions were implemented for relatively simple operations such as element-wise operations. For RNNLM computations on CPUs such as output layer computations, we used EIGEN which is a C++ based linear algebra library.

\subsection{Results}
\label{ssec:results}

\begin{figure}[h]
	\centering
	\centerline{\includegraphics[width=8.5cm]{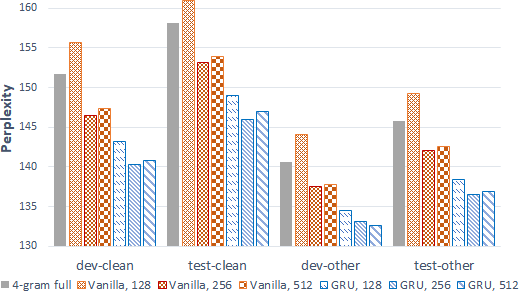}}
	\caption{Perplexities depending on LM types.}
	\label{fig:perplexities}
\end{figure}

In our experiments, LibriSpeech's development and test cases were used for evaluations.
The performance of different LMs measured in terms of perplexity is shown in Figure~\ref{fig:perplexities}.
The term ``other" in the evaluation cases means the speech data sets were recorded in noisy environments.
As can be seen in Figure~\ref{fig:perplexities}, the vanilla-RNNLM of size 128 showed the worst accuracies over all the sets and was even worse than that of the 4-gram LM.
The accuracy of vanilla-RNNLM improved dramatically for a hidden layer size of 256 and showed the lowest perplexities, but still worse than a 128-size GRU-RNNLM.
Perplexities of GRU-RNNLMs were dropped by 7.81, 10.10, and 9.75 absolute (averaged over all four cases) for model sizes of 128, 256, and 512, respectively, as compared to the perplexity of the 4-gram LM.
In all tasks except for ``dev-other," the GRU-RNNLM size of 256 showed the lowest LM perplexities.

Table~\ref{tab:performance} shows the word error rate (WER) and the real-time factor (RTF) for the proposed methods for accelerating the online RNNLM rescoring.
All decoding options otherwise mentioned in Table~\ref{tab:performance} are same for all the methods being compared.
The meanings of values in the column ``Precision" are the same as Table~\ref{tab:precision}.
Regarding recognition accuracies, the average WER of the baseline system was improved by 3.39\% relatively than that of the 4-gram LM based system as can be observed from the first two rows of Table~\ref{tab:performance}.
As expected in Section 3, caching quantized history vectors rounded off in the first and the second decimal points did not show noticeable improvement in recognition speed compared to the baseline system.
However, the proposed quantization strategy of caching only signs of the history vectors was 1.23 times faster compared to the baseline system without any accuracy degradations.

As shown in the fifth and sixth rows of Table~\ref{tab:performance}, with the proposed GPU parallelization method, even one thread was 1.43 times faster on an average than the fastest CPU based system (sign).
The recognition speed improves further with the use of multiple GPUs.
In particular, when the number of GPUs increased to two, the speed was significantly improved, which was 1.61 times faster than a single GPU-based system.
When three GPUs were utilized, we attained real-time speech recognition over all the test cases.
Finally, the RNNLM-based ASR system with four GPUs has shown the fastest average recognition speed of 0.72 RTF.
It was three times faster than the fastest CPU-based system and four times faster than the baseline system.

\section{conclusion}
\label{sec:conclusion}
We devised a faster RNNLM based on-the-fly rescoring on both CPU and GPU platforms by introducing a lossy compression strategy of history vectors and the novel hybrid parallelization method.
As cache hit ratios got higher by lowering decimal precisions of the vectors, speech recognition was speeded up by 1.23 times.
Although it was not a significant improvement, the fact that recognition rates were not affected even if each dimension of the history vectors was stored by one bit representing the sign seemed to provide a clue to the efficient compression way of embedding vectors while minimizing the loss of their information.
Finally, with the CPU-GPU hybrid parallelization method, the decoding speed over all the cases has fallen within real-time.

\bibliographystyle{IEEEbib}
\bibliography{refs}

\end{document}